\documentclass{article}

\usepackage{arxiv}

\usepackage[utf8]{inputenc} % allow utf-8 input
\usepackage[T1]{fontenc}    % use 8-bit T1 fonts
\usepackage{hyperref}       % hyperlinks
\usepackage{url}            % simple URL typesetting
\usepackage{booktabs}       % professional-quality tables
\usepackage{amsfonts}       % blackboard math symbols
\usepackage{nicefrac}       % compact symbols for 1/2, etc.
\usepackage{microtype}      % microtypography
\usepackage{cleveref}       % smart cross-referencing
\usepackage{lipsum}         % Can be removed after putting your text content
\usepackage{graphicx}
\usepackage{natbib}
\usepackage{doi}

\usepackage{booktabs}
\usepackage{graphicx}
\usepackage[table,xcdraw]{xcolor}

\title{SPOTS-10: Animal Pattern Benchmark Dataset for Machine Learning Algorithms}

\author{ \href{https://orcid.org/0000-0003-2307-2720}{\includegraphics[scale=0.06]{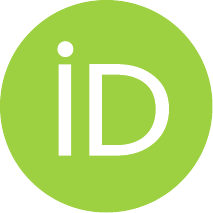}\hspace{1mm}John ~Atanbori}\thanks{Use footnote for providing further
		information about author (webpage, alternative
		address)---\emph{not} for acknowledging funding agencies.} \\
	University of Lincoln\\
	School of Engineering and Physical Sciences\\
	Brayford Way, Brayford Pool \\
	Lincoln, LN6 7TS \\
	United Kingdom\\
	\texttt{jatanbori@lincoln.ac.uk} \\
}

\renewcommand{\shorttitle}{\textit{arXiv} Template}

\hypersetup{
pdftitle={SPOTS-10: Animal Pattern Benchmark Dataset for Machine Learning Algorithms},
pdfsubject={},
pdfauthor={John ~Atanbori},
pdfkeywords={},
}

\begin{document}
\maketitle

\begin{abstract}
	Recognising animals based on distinctive body patterns, such as stripes, spots, or other markings, in night images is a complex task in computer vision. Existing methods for detecting animals in images often rely on color information, which is not always available in night images, posing a challenge for pattern recognition in such conditions. Nevertheless, recognition at nighttime is essential for most wildlife, biodiversity, and conservation applications. The SPOTS-10 dataset was created to address this challenge and to provide a resource for evaluating machine learning algorithms in situ. This dataset is an extensive collection of grayscale images showcasing diverse patterns found in ten animal species. Specifically, SPOTS-10 contains 50,000 32 × 32 grayscale images, divided into ten categories, with 5,000 images per category. The training set comprises 40,000 images, while the test set contains 10,000 images. The SPOTS-10 dataset is freely available on the project GitHub page:  \href{https://github.com/Amotica/SPOTS-10.git}{https://github.com/Amotica/SPOTS-10.git} by cloning the repository.
\end{abstract}

% keywords can be removed

\section{Introduction}
\label{intro}
Detecting animals from their patterns, such as stripes, spots, or other markings, in grayscale camera trap images is a challenging computer vision task that is crucial for various environmental and ecological applications. This task is even more difficult due to the natural environments where these camera traps are located, including dense foliage and changing light conditions. The SPOTS-10 dataset is created to help tackle these challenges by providing extensive benchmark data for developing and evaluating machine learning algorithms for \textbf{pattern-based} classification.

\begin{figure}[hbt!]
	\centering
	\includegraphics[width=1.0\textwidth]{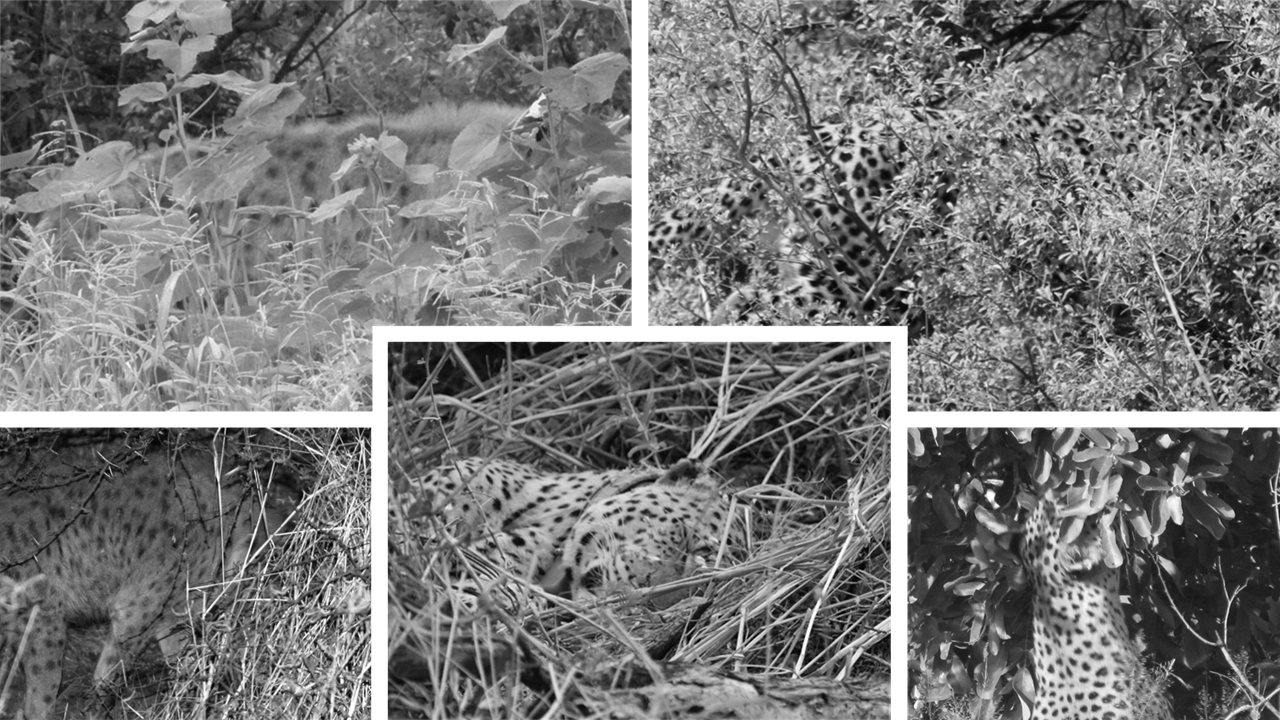}
	\caption{Sample images of a leopard and a hyena are partially occluded by plants. This is a common situation in animal species datasets, as these animals often hide to ambush their prey. Therefore, the only visible features for identification are partial distinct markings that can be spotted through the bushes.}
	\label{fig_animal_partially_visible}
\end{figure}

Camera traps have become an invaluable tool in wildlife research and conservation efforts. These devices are strategically placed in wildlife habitats to capture images of animals in their natural settings \cite{marcus2011quantifying}. They offer insights into animal behaviour, population dynamics, and ecosystem health without human interference. Identifying animals in these images can be highly challenging due to several factors: 1) Animals are often partially obscured, for example, by vegetation (see Figure \ref{fig_animal_partially_visible}). 2) The images are sometimes captured in grayscale, particularly during nocturnal hours when some animals are active, which presents some challenges to computer vision approaches. However, \cite{villa2017towards} and \cite{binta2023animal} have demonstrated that species could be detected in such images by learning complex patterns using convolutional neural networks (CNNs).

Patterns such as stripes, spots, and other unique markings are crucial for identifying many animal species. For instance, the spots on a leopard or the stripes on a tiger are distinctive features of distinguishing species. However, detecting these patterns in grayscale images and when animals are heavily occluded by forest cover presents significant technical challenges. The SPOTS-10 dataset provides extensive benchmark data to develop and test machine learning (ML) algorithms for this purpose, and to also quickly benchmark ML algorithms for general classification tasks. To achieve this aim, we have collected large numbers of images for ten species of animals, extracted patches of distinctive markings from the images, and applied preprocessing algorithms to the patches to simulate night vision.  

\section{The SPOTS-10 Dataset}
\label{spots_10_dataset}

\begin{figure}[hbt!]
	\centering
	\includegraphics[width=1.0\textwidth]{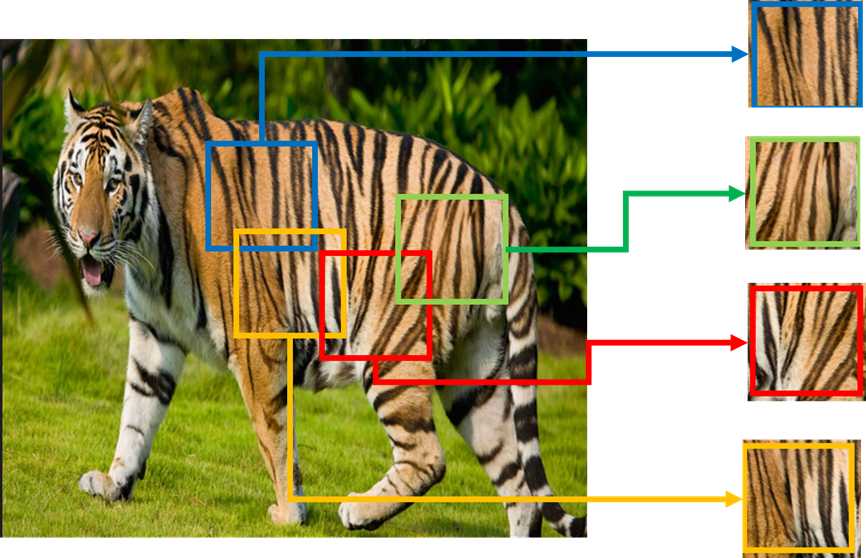}
	\caption{Randomly create non-fully overlapping 90x90 patches from the Tiger category, with a maximum of six patches per image.}
	\label{fig_patches_creation}
\end{figure}

Most of the images in this dataset were collected over six months period and obtained by searching the web for images of various animal species with spots, stripes, and other distinctive patterns. The search queries mainly included the animals' common and scientific names using search engines like Google and Flickr and retaining approximately the first 2,000 search results for each search term with the license "CC BY", "CC BY-SA", and "CDLA-Permissive". After gathering all the images for a specific search term, non-natural (synthetic) images were removed together with perfect duplicates. The images were then categorized using the broad common names of the species. Image patches of size 90 x 90, downsized later with the pipeline to 32 x 32, were extracted from the downloaded images,  representing species patches with stripes, spots, or other distinct markings. In the visualization of the process shown in Figure \ref{fig_patches_creation}, partial overlaps were  allowed while avoiding complete overlaps. Some dataset categories with fewer samples were supplemented by incorporating images from additional sources. For instance, Whale Shark images were obtained from \cite{holmberg2009estimating}, zebra and giraffe images from \cite{parham2017animal}, and some hyena and leopard images from \cite{botswana2022panthera}.

\begin{figure}[hbt!]
	\centering
	\includegraphics[width=1.0\textwidth]{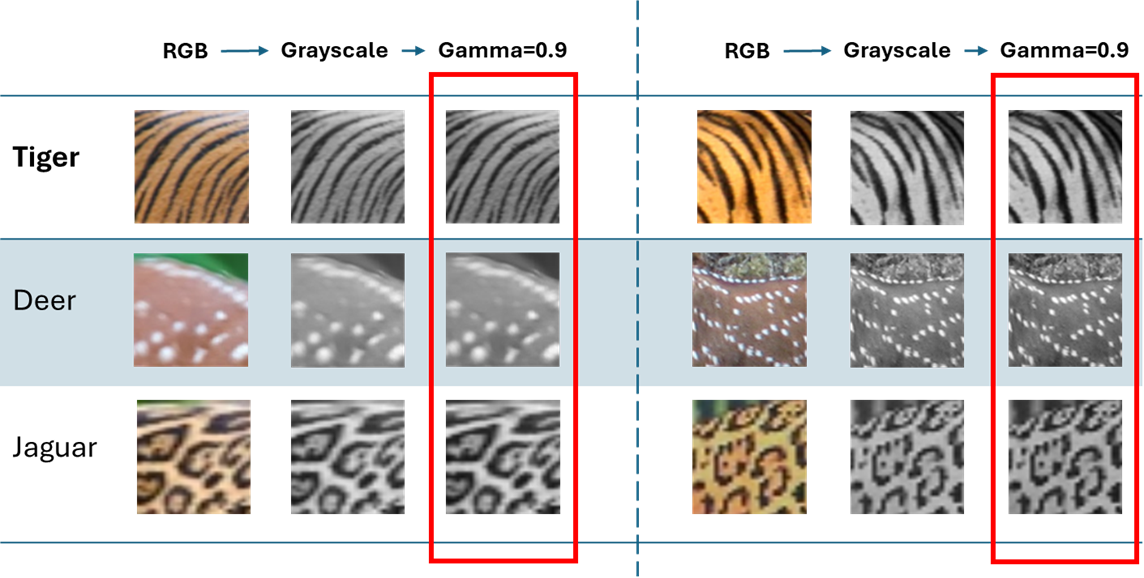}
	\caption{Visualization of the conversion pipeline: RGB image converted to grayscale, followed by the application of inverse gamma correction of 0.9.}
	\label{fig_conversion_pipeline}
\end{figure}

The	90 x 90 patches representing spots, stripes, or distinctive markings of the various categories were preprocessed by converting them into grayscale to mimic camera trap night images, and their visualization is in Figure \ref{fig_conversion_pipeline}, and below is the conversion pipeline to achieve this:

\begin{itemize}
	\item Read 90x90 image patches for each category.
	\item Convert image patches to greyscale.
	\item Apply inverse gamma correction, which is done to enhance the visibility of details in dark regions of the image patches without overexposing bright areas.
	\begin{itemize} 
		\item Normalize the image to the range [0, 1]
		\item Apply the inverse gamma correction of 0.9 to all image patches.
	\end{itemize}
	\item Scale back to range [0, 255]  and convert the image to 8-bit grayscale pixels.
	\item Save the image patches in the appropriate output category as PNG images.
\end{itemize}

\begin{figure}[hbt!]
	\centering
	\includegraphics[width=0.7\textwidth]{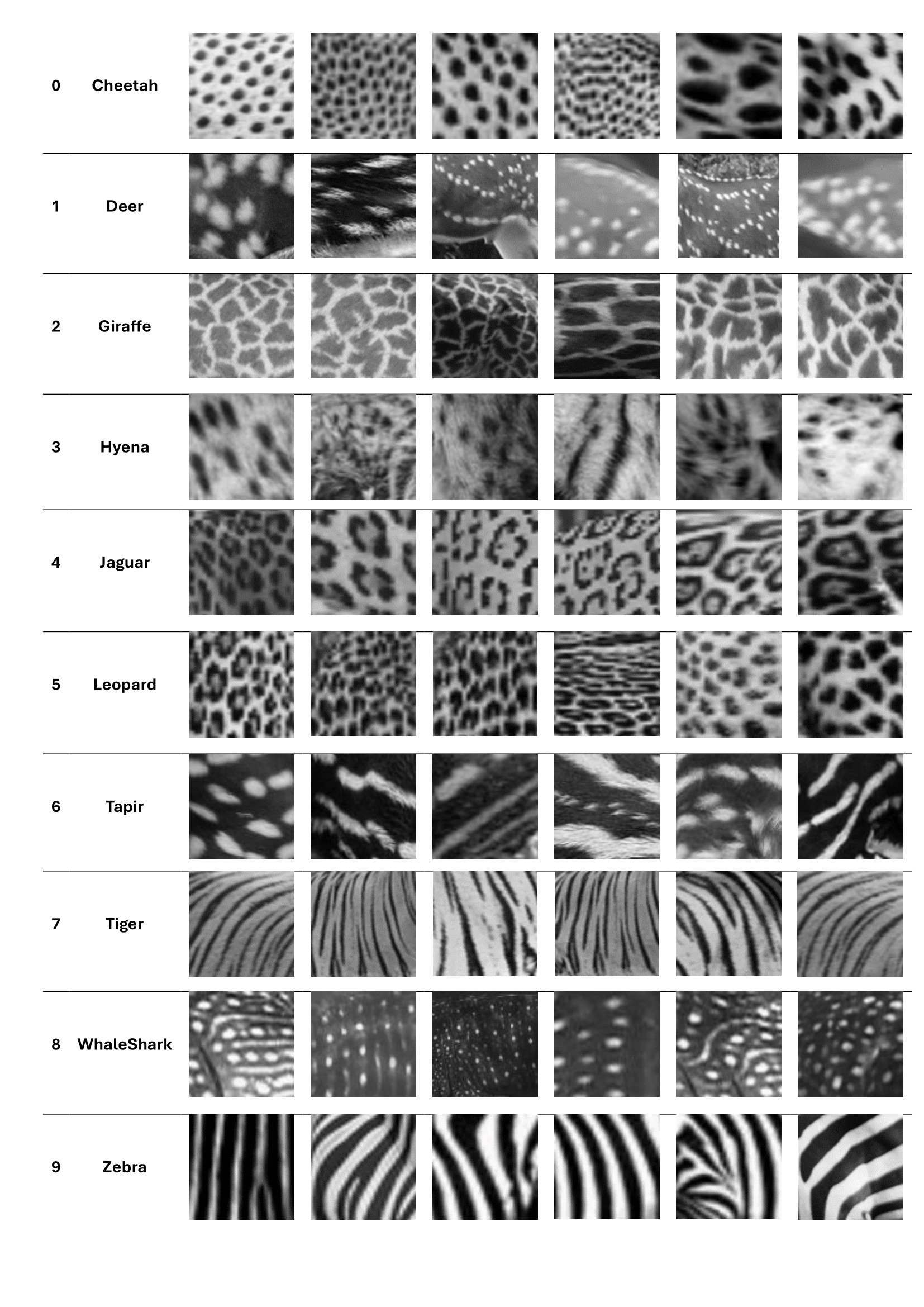}
	\caption{Shows varying samples from the complete dataset for each of the ten classes, including the class ID (label ID) and class names.}
	\label{fig_complete_dataset}
\end{figure}

The SPOT-10 dataset, used in the baseline experiments reported in section \ref{experiments}, was obtained after applying the conversion pipeline. A sample visualization of the SPOT-10 dataset is presented in Figure \ref{fig_complete_dataset}, which summarises all categories with samples for each category showing the variation in the dataset. The final dataset is divided into training and testing sets to perform baseline experiments. The training set consists of 4,000 randomly selected patch examples from each category, while the test set consists of 1,000 patches per category. 
The patches in the training and test sets are disjoint, meaning there is no overlap between them.

\begin{table}[]
	\caption{Shows details of the files in the SPOT-10 dataset, including name, description, sample size, and size of image patches.}
	\resizebox{\textwidth}{!}{%
		\begin{tabular}{@{}|l|l|l|l|@{}}
			\toprule
			\rowcolor[HTML]{C0C0C0} 
			\textbf{Name}              & \textbf{Description}                     & \textbf{(Samples, Dimension)} & \textbf{Size} \\ \midrule
			train\-images\-idx3\-ubyte.gz & Images in the training set               & (40000, 32, 32)               & 8.87 Mbytes   \\ \midrule
			train\-labels\-idx1\-ubyte.gz & Labels or categories of the training set & (40000, 1)                    & 234 bytes     \\ \midrule
			t10k\-images\-idx3\-ubyte.gz  & Images in the testing set                & (10000, 32, 32)               & 35.4 MBytes   \\ \midrule
			t10k\-labels\-idx1\-ubyte.gz  & Labels or categories of the training set & (10000, 1)                    & 122 bytes     \\ \bottomrule
		\end{tabular}%
	}
	\label{tbl_files_in_spot_10_dataset}
\end{table}

Images and labels are stored in the same format as the MNIST and Fashion-MNIST datasets using the idx3-ubyte format, designed for storing vectors and multidimensional matrices. Before converting to this format, the images were all resized to 32 x 32, and the resultant idx3-ubyte data were compressed using gzip before storage.  The summary of the resulting dataset files is listed in Table \ref{tbl_files_in_spot_10_dataset}.

\section{Baseline Experiments}
\label{experiments}

% \usepackage{graphicx}
\begin{table}[hbt!]
	\caption{Shows the results of benchmark experiments performed with various distiller models, reporting the accuracy and size of each model alongside the base teacher model on the ImageNet dataset to aid in the evaluation of the distiller models.}
	\resizebox{\textwidth}{!}{%
		\begin{tabular}{|lcc|lcc|}
			\hline
			\multicolumn{3}{|c|}{\textbf{Distiller Models on SPOTS-10}}                                                                                             & \multicolumn{3}{c|}{\textbf{Original Models on ImageNet}}                                                                                              \\ \hline
			\rowcolor[HTML]{C0C0C0} 
			\multicolumn{1}{|l|}{\cellcolor[HTML]{C0C0C0}\textbf{Model Name}} & \multicolumn{1}{c|}{\cellcolor[HTML]{C0C0C0}\textbf{Accuracy}} & \textbf{Size (MB)} & \multicolumn{1}{l|}{\cellcolor[HTML]{C0C0C0}\textbf{Model Name}} & \multicolumn{1}{c|}{\cellcolor[HTML]{C0C0C0}\textbf{Accuracy}} & \textbf{Size (MB)} \\ \hline
			\multicolumn{1}{|l|}{DenseNet121 Distiller}                       & \multicolumn{1}{c|}{81.84\%}                                   & 0.07               & \multicolumn{1}{l|}{DenseNet121}                                 & \multicolumn{1}{c|}{75.00\%}                                   & 33                 \\ \hline
			\multicolumn{1}{|l|}{ResNet101V2 Distiller}                       & \multicolumn{1}{c|}{80.29\%}                                   & 0.07               & \multicolumn{1}{l|}{ResNet101V2}                                 & \multicolumn{1}{c|}{77.20\%}                                   & 171                \\ \hline
			\multicolumn{1}{|l|}{ResNet50V2 Distiller}                        & \multicolumn{1}{c|}{79.03\%}                                   & 0.07               & \multicolumn{1}{l|}{ResNet50V2}                                  & \multicolumn{1}{c|}{76.00\%}                                   & 98                 \\ \hline
			\multicolumn{1}{|l|}{MobileNet Distiller}                         & \multicolumn{1}{c|}{78.26\%}                                   & 0.07               & \multicolumn{1}{l|}{MobileNet}                                   & \multicolumn{1}{c|}{70.40\%}                                   & 16                 \\ \hline
			\multicolumn{1}{|l|}{MobileNetV3Small Distiller}                  & \multicolumn{1}{c|}{78.04\%}                                   & 0.07               & \multicolumn{1}{l|}{MobileNetV3Small}                            & \multicolumn{1}{c|}{68.10\%}                                   & 2.9                \\ \hline
			\multicolumn{1}{|l|}{ResNet101 Distiller}                         & \multicolumn{1}{c|}{78.01\%}                                   & 0.07               & \multicolumn{1}{l|}{ResNet101}                                   & \multicolumn{1}{c|}{76.40\%}                                   & 171                \\ \hline
			\multicolumn{1}{|l|}{MobileNetV3Large Distiller}                  & \multicolumn{1}{c|}{77.88\%}                                   & 0.07               & \multicolumn{1}{l|}{MobileNetV3Large}                            & \multicolumn{1}{c|}{75.60\%}                                   & 5.4                \\ \hline
			\multicolumn{1}{|l|}{NASNetMobile Distiller}                      & \multicolumn{1}{c|}{77.75\%}                                   & 0.07               & \multicolumn{1}{l|}{NASNetMobile}                                & \multicolumn{1}{c|}{74.40\%}                                   & 23                 \\ \hline
			\multicolumn{1}{|l|}{MobileNetV2 Distiller}                       & \multicolumn{1}{c|}{77.53\%}                                   & 0.07               & \multicolumn{1}{l|}{MobileNetV2}                                 & \multicolumn{1}{c|}{71.30\%}                                   & 14                 \\ \hline
			\multicolumn{1}{|l|}{ResNet50 Distiller}                          & \multicolumn{1}{c|}{77.45\%}                                   & 0.07               & \multicolumn{1}{l|}{ResNet50}                                    & \multicolumn{1}{c|}{74.90\%}                                   & 98                 \\ \hline
		\end{tabular}%
	}
	\label{tbl_benchmark_results}
\end{table}

To benchmark this new dataset the knowledge distillation method from \cite{hinton2015distilling} was employed, but without ensembles. In this approach, a small model, known as the student model, was developed and trained to emulate the performance of a large pre-trained teacher model. The aim was to transfer the knowledge from the teacher model to the student model by optimizing a loss function that aligns the student model's outputs with both the softened logits of the teacher model and the ground-truth labels \cite{}. The teacher models used were pre-trained using the following ten model architectures: \textbf{ResNet50} \cite{he2016deep}, \textbf{MobileNet} \cite{howard2017mobilenets}, \textbf{MobileNetV2} \cite{sandler2018mobilenetv2}, \textbf{DenseNet121} \cite{huang2017densely}, \textbf{NASNetMobile} \cite{zoph2018learning}, \textbf{MobileNetV3 Large} \cite{howard2019searching}, \textbf{MobileNetV3 Small} \cite{howard2019searching}, \textbf{ResNet50V2} \cite{huang2017densely}, \textbf{ResNet101V2} \cite{huang2017densely}, and \textbf{ResNet101} \cite{he2016deep}. Detailed implementations of these models can be found in Tensorflow Applications at \href{https://www.tensorflow.org/api_docs/python/tf/keras/applications}{https://www.tensorflow.org/api\_docs/python/tf/keras/applications}, and citations to the original papers that implemented the algorithms are provided for reference. The weights of these models were those trained on the ImageNet dataset.

\begin{figure}[hbt!]
	\centering
	\includegraphics[width=0.7\textwidth]{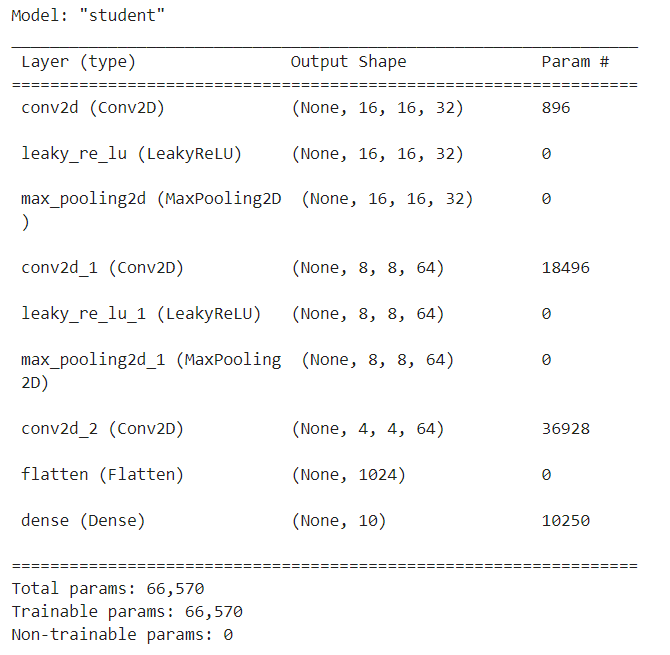}
	\caption{Shows the student model architecture used to emulate the behaviour of the base teacher models.}
	\label{fig_student_model}
\end{figure}

The student model architecture (Figure \ref{fig_student_model}) consists of an input layer followed by three convolutional blocks. Each block includes a Conv2D layer (32 filters in the first, 64 in the second and third), a LeakyReLU activation with an alpha of 0.2, and a MaxPooling2D layer with a pool size of (2, 2) and strides of (1, 1). After the convolutional blocks, a Flatten layer converts the 2D output to a 1D vector, followed by a Dense layer with units equal to the number of classes. The model is designed to be efficient and straightforward for processing and classifying input data.

The distillation process involves using a Distiller model, which pairs a student model with a teacher model. The distiller is compiled with an Adam optimizer, sparse categorical accuracy metric, and loss functions for both the student model (sparse categorical cross-entropy) and the distillation process (Kullback-Leibler(KL) divergence), with an alpha value of 0.1 and a temperature of 10. Two callbacks are defined: ReduceLROnPlateau to reduce the learning rate by a factor of 0.5 if the validation accuracy does not improve for 5 epochs, and ModelCheckpoint to save the best model based on validation accuracy. The distiller is then trained on the training data  for 100 epochs with a batch size of 32 and incorporating the defined callbacks to adjust the learning rate and save the best model during training.

The detailed implementation for benchmarking is available on the project's \href{https://github.com/Amotica/SPOTS-10.git}{ GitHub page}. All teacher models are implemented using Keras applications in Tensorflow to ensure accurate testing and reproducibility. The same student model is used in each benchmark, paired with different teacher models, and each benchmark model is named after the corresponding teacher model. For instance, the benchmark using the MobileNet architecture is referred to as "MobileNet Distiller" in this paper. 

Baseline experiments carried out are presented in Table \ref{tbl_benchmark_results}, which shows the distiller models' accuracy and size on SPOTS-10 as well as the teacher models' accuracy and size on ImageNet. The teachers' accuracies on ImageNet were presented to demonstrate the student models' level of emulation for comparison. ResNet101V2, ResNet101, ResNet50V2, MobileNetV3Large, and DenseNet121 were among the top five performers in terms of accuracy on ImageNet, which reflected on the distillation process of the student models. This process resulted in DenseNet121 Distiller, ResNet101V2 Distiller, ResNet50V2 Distiller, and ResNet101 Distiller being among the top five performers in accuracy. MobileNetV3Large was among the lower-performing models among the distillers, even though the accuracy attained by this model is still comparable to the others.

\section{Conclusion}
\label{conclusion}

A new dataset, SPOTS-10, is introduced in the paper to benchmark machine-learning algorithms for pattern recognition in night images of animal species. The dataset contains 50,000 32 × 32 grayscale images, divided into ten categories, with 5,000 images per category. The training set comprises 40,000 images, while the test set contains 10,000 images. The SPOTS-10 dataset is freely available on the project's GitHub page: \href{https://github.com/Amotica/SPOTS-10.git}{https://github.com/Amotica/SPOTS-10.git} by cloning the repository. Initial benchmarking has also been provided with ten CNN architectures focusing on the knowledge distillation method. A small model, the student model, was developed and trained to emulate the performance of a large pre-trained teacher model, transferring the knowledge from the teacher model to the student model.

\bibliographystyle{unsrtnat}
\bibliography{references} 

\end{document}